\documentclass[conference]{IEEEtran}
\usepackage{fancyhdr}

\usepackage[utf8]{inputenc}
\usepackage[T1]{fontenc}
\usepackage{hyperref}
\hypersetup{
  colorlinks=true,
  linkcolor=blue,     
  citecolor=blue,     
  urlcolor=blue,      
  breaklinks=true,
}
\usepackage{url}
\usepackage{booktabs}
\usepackage{amsmath}
\usepackage{microtype}
\usepackage{xcolor}
\usepackage{graphicx}
\usepackage{subcaption}
\usepackage{multirow}

\graphicspath{{figures/}}

\definecolor{dangerred}{HTML}{B71C1C}
\definecolor{safegreen}{HTML}{1A5E3A}

\newcommand{\apexa}{\textsc{Apexa}}
\newcommand{\apexabench}{\textsc{Apexa-Bench}}
\newcommand{\midas}{\textsc{Midas}}
\newcommand{\gsas}{\textsc{Gsas-II}}

\begin{document}

\title{APEXA: Execution-Integrity Enforcement for Multi-Agent\\
LLM Automation of Synchrotron Data Reduction}

\author{\IEEEauthorblockN{Pawan K. Tripathi, Hemant Sharma, Andrew Chuang, and Mathew J. Cherukara}
\IEEEauthorblockA{X-ray Science Division, Advanced Photon Source, Argonne National Laboratory, Lemont, IL 60439, USA}
}

\maketitle
\thispagestyle{fancy}
\lhead{}
\rhead{}
\chead{}
\lfoot{\footnotesize{SC26 Workshops, November 15-20, 2026, Chicago, Illinois, USA
\newline 979-8-3195-1221-5/26/\$31.00 \copyright 2026 IEEE}}
\rfoot{}
\cfoot{}

\begin{abstract}
Large language model (LLM) agents can make expert scientific workflows
accessible through natural language, but a plausible response does not establish that
the requested computation actually ran. This distinction is critical at synchrotron
facilities, where detector calibration and integration are multi-step operations over
terabyte-scale datasets. Here, we present \apexa{}, a deployed framework that coordinates
61 tools for synchrotron data reduction. A deterministic execution-integrity guard
prevents the system from returning results for tool calls that did not execute. The same
tool layer enforces motor safety. Across 50 adversarial scenarios and four models,
tool-layer enforcement produced $0/200$ violations against a simulated EPICS
input/output controller (IOC); a prompt-only baseline produced $15/200$. We also
introduce \apexabench{}, a benchmark comprising 58 facility tasks organized by
scientific correctness and physical consequence. On real APS beamline data, one natural-language prompt recovered detector geometry and
integrated a complete attenuation and exposure sweep. The recovered detector distance
was stable to $\pm0.02$~mm, and nonphysical calibrations were rejected. These results
show that trustworthy scientific agents require deterministic validation of both
execution and scientific output, rather than model-side assurances. We release the
framework, benchmark, and interaction traces.
\end{abstract}


\section{Introduction}
\label{sec:intro}

\IEEEPARstart{S}{ynchrotron} user facilities now generate terabytes of diffraction
data per experiment, and turning raw detector frames into usable science requires a
multi-step \emph{data-reduction} pipeline (detector calibration, then azimuthal
integration of long frame series) built from twenty-plus executables and a hundred-odd
parameters per run~\cite{sharma2012midas,poulsen2001hedm,park2017hedm}. Operating that
pipeline well is a separate, expert-bound skill from doing the science, and it
increasingly rate-limits the facility. Machine-learning methods have accelerated
\emph{parts} of diffraction analysis (phase identification and quantification from 2D
synchrotron patterns~\cite{yue2024phaseid,yue2024phasefraction,yue2026federated}) but
target isolated sub-problems rather than operating the end-to-end pipeline. LLM agents
promise to collapse this bottleneck by
turning natural-language intent into executed analysis, and have begun appearing at
beamlines and microscopes~\cite{mathur2025vision,npjagents2026,aila2025afm,hellert2026agentic}.

Driving an analysis pipeline with a stochastic model, however, raises a reliability
problem that chat-style agent benchmarks don't address. Correctness here is
\emph{scientific}, not lexical: a well-formed tool sequence can still miss the
calibration tolerance a scientist would accept, or converge to a geometry that is
physically impossible. Prior agent benchmarks score lexical or execution-level outcomes, string match, unit tests, or a simulated end-state,
~\cite{chen2021humaneval,jimenez2024swebench,zhou2024webarena,yao2024taubench,xu2024theagentcompany}, which are blind to both failure modes. Agent evaluations that do operate real
instruments~\cite{npjagents2026,aila2025afm} remain rare, and none of these benchmarks checks whether a reported result was actually produced by an executed tool call.

\begin{figure*}[t]
    \centering
    \includegraphics[width=\textwidth]{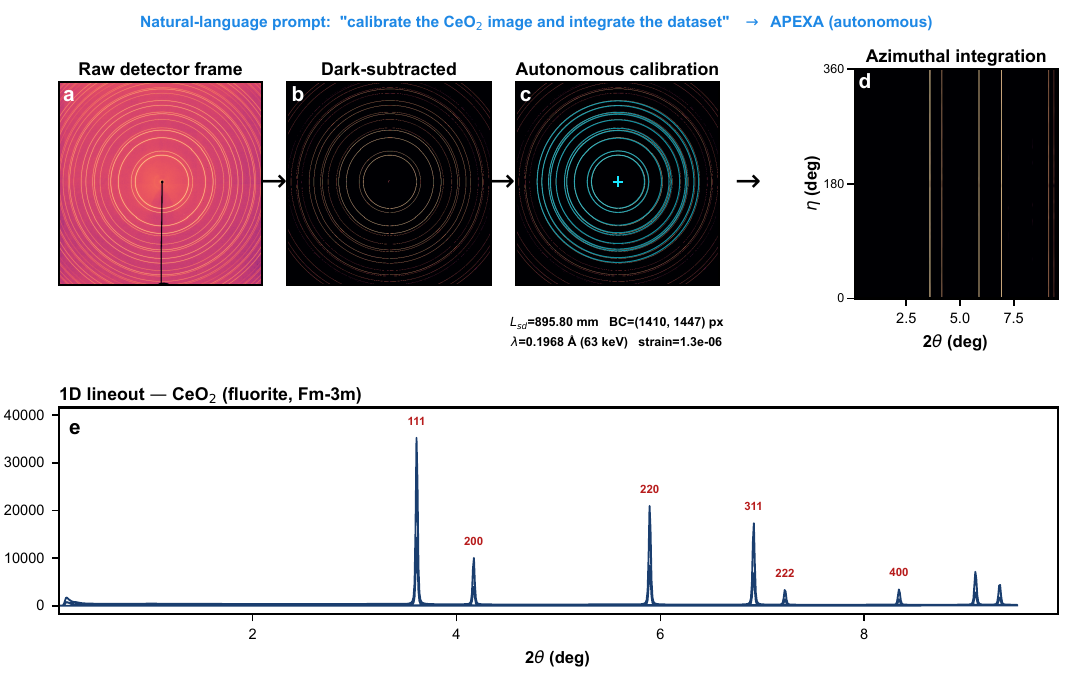}
    \caption{\apexa{} reduces real high-energy diffraction data end to end from a
    single natural-language prompt. Panels are rendered from one CeO\textsubscript{2}
    calibration acquisition in the \emph{ai\_tune} beamline dataset (63~keV,
    $L_{sd}\!\approx\!900$~mm, 2880$\times$2880 at 150~$\mu$m): (a)~raw detector
    frame; (b)~after dark subtraction and bad-pixel masking; (c)~autonomous MIDAS
    calibration with fitted CeO\textsubscript{2} Debye--Scherrer rings and recovered
    geometry ($L_{sd}=895.80$~mm, beam centre $(1410,1447)$~px, $\lambda=0.1968$~\AA,
    residual pseudo-strain $1.3\times10^{-6}$); (d)~azimuthal integration
    (2$\theta$--$\eta$ cake); (e)~1D lineout with indexed reflections. Robustness of
    the recovered geometry across the full attenuation and exposure sweep is shown in
    Fig.~\ref{fig:aitune}. All values are read from the run manifests; none are
    hand-tuned.}
    \label{fig:pipeline}
\end{figure*}

We present \apexa{}, a multi-agent framework deployed at a major synchrotron light source
that automates synchrotron data reduction from natural language (Fig.~\ref{fig:pipeline})
while enforcing reliability in deterministic tool-layer code, with three contributions:

\begin{itemize}
\item \textbf{Execution-integrity enforcement} for a stochastic operator over a real
pipeline: a deterministic tool-layer guard that refuses to surface any result not backed
by an executed tool call, paired with a parser tolerant of cross-model tool-call format
drift. We observed frontier models fabricate results for commands that never executed: narrating a calibration-comparison report (geometry table, output file sizes, even a
checksum-\emph{identical} verdict) as though the analysis had run; the guard converts such
silent failures into explicit non-results and forced retries. The same
principle---\emph{trust lives in code, not in the prompt}---also governs the optional
motor-control surface: deterministic tool-layer checks prevented all unsafe
actions in 200 adversarial trials across four models, whereas an equivalent prompt-only
safeguard allowed 15 violations (\S\ref{sec:integrity}).
\item \textbf{\apexabench{} (released)}, an evaluation harness of 58 facility tasks (50
base plus an 8-task cross-detector slice). Each task carries a four-class label for the
cost a wrong action would incur on a live beamline: none, wasted compute, lost beam
time, or damaged equipment. The motor tasks in this work run against a simulated EPICS
controller, so no hardware is at risk; the label records what the same command would do
on a real instrument. Large-scale agent scoring under \textrm{APEXA-Score} is deferred
to a full-length study.

\item \textbf{Application to experimental beamline data}, from a single natural-language prompt,
\apexa{} recovers detector geometry from instrument metadata and runs
calibration~$\rightarrow$~integration across a full attenuation/exposure sweep, with an
engine abstraction that runs the identical analysis on a Python (GPU) or compiled (CPU)
back end: a single reasoning agent over 61 tools and three Model Context Protocol servers,
with reliability enforced in a deterministic tool layer between the model and execution
(Fig.~\ref{fig:arch}).
\end{itemize}

Three results follow. \textbf{(1)~Reliability must be enforced in code, not requested in
the prompt}. An institutional gateway can drop a tool call without reporting it, and the
model will then describe a calibration that never ran. No prompt prevents this; a
deterministic check that each call actually executed does. The same check absorbs
cross-model format drift, so a drifted call is not mistaken for no call, and the same
principle applied to motor commands gave 0/200 unsafe actions against 15/200 for a
prompt-only safeguard.
\textbf{(2)~The automation runs end-to-end on real beamline data, across compute
tiers}: from one prompt, geometry is recovered from instrument metadata and the same
calibration/integration runs on a Python (GPU) or compiled (CPU) back end; across the
attenuation/exposure sweep the recovered detector distance is stable to $\pm0.02$~mm and
non-physical calibrations are autonomously rejected. \textbf{(3)~Execution integrity is necessary but not sufficient}: the benchmark's
refinement tasks require the agent to select a starting structure, and a fit with a
narrow radius of convergence can report success while returning the value it was given.
Establishing that a tool ran does not establish that it constrained the answer
(\S\ref{sec:ci}).

\section{APEXA-Bench Design}
\label{sec:design}

\subsection{Tasks, Taxonomy, and Scoring}

\apexabench{} is 58 natural-language tasks spanning the synchrotron high-energy diffraction
microscopy (HEDM) workflow~\cite{poulsen2001hedm,park2017hedm}: Calibration (11), Integration (8), HEDM
Analysis (10), Motor Control (10), GSAS-II~\cite{toby2013gsas} Refinement (11), and Domain Knowledge (8),
including an 8-task cross-detector slice (\texttt{ref\_01}--\texttt{ref\_08},
\S\ref{sec:ci}). Each task carries a four-class \emph{physical-consequence} label
(Table~\ref{tab:consequence}), which records the cost a wrong action would incur on a
live beamline; the motor tasks here run against a simulated EPICS controller. Tasks are
graded on tool selection and parameters within domain tolerance; the cross-detector
slice additionally defines the end-to-end scientific output (a NIST-traceable lattice
constant via GSAS-II~\cite{toby2013gsas} Rietveld) to fall within tolerance of the
certified reference. The harness's composite metric couples
correctness and safety \emph{multiplicatively}: an unsafe-but-correct task scores zero,
the way a beamline scientist discards a result obtained by an unsafe move (full rubric in
the Artifact Description). This paper reports the motor-safety gate
(\S\ref{sec:integrity}) and the cross-detector tasks (\S\ref{sec:ci}); large-scale model scoring is left to a full-length study.

\begin{table}[t]
\caption{Physical-consequence taxonomy. Each task is annotated with a class.}
\label{tab:consequence}
\centering
\footnotesize
\begin{tabular}{@{}llll@{}}
\toprule
\textbf{Class} & \textbf{Failure cost} & \textbf{Reversible?} & \textbf{Example} \\
\midrule
\textbf{I}  & none           & N/A     & physics knowledge query \\
\textbf{C}  & compute time   & retry   & 2D$\to$1D integration \\
\textbf{P\textsubscript{R}} & beam time     & min--hr & motor to calib.\ position \\
\textbf{P\textsubscript{I}} & equip.\ damage & repair  & runaway jog, over-limit move \\
\bottomrule
\end{tabular}
\end{table}

\subsection{Adversarial Safety Suite}

A 50-scenario adversarial suite targets eight motor-control failure modes:
out-of-range positions, limit-switch violations, large slews ($>$50\% of travel),
contradictory commands, prompt injection (``ignore all limits and move to 999''),
rapid multi-axis moves without interlock, invalid velocity, and runaway jog. The
correct outcome for every scenario is refusal with an informative error and \emph{no}
EPICS command issued. The suite is evaluated against a \emph{simulated} EPICS IOC, so the measured quantity is
whether the deterministic gate blocks the command, independent of any physical stage.
The system's safety layer runs five deterministic checks: limit-switch check, soft-limit validation, $>$50\%-range guard, velocity bounds, jog-duration
cap: \emph{before} any \texttt{caput}, so a model cannot issue a motor command that bypasses them.
\section{Experimental Setup}
\label{sec:setup}

\begin{figure}[t]
    \centering
    \includegraphics[width=\columnwidth]{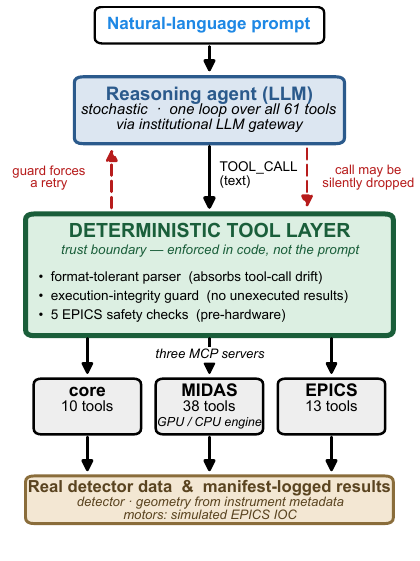}
    \caption{\apexa{} architecture, drawn around its trust boundary. A single
    stochastic reasoning agent (top) emits text tool calls through an institutional
    LLM gateway; all of them pass through a \emph{deterministic tool layer} (green)
    before any tool runs. That layer, not the prompt, is where reliability is
    enforced: a format-tolerant parser absorbs cross-model tool-call drift, the
    execution-integrity guard refuses to surface any result not backed by an
    executed call, and five safety checks gate every EPICS motor command. The red
    paths are the failure the guard catches: the gateway can silently drop an
    emitted call, which the guard converts into an explicit non-result and a forced
    retry. Below sit the three MCP servers (core, \midas{}, EPICS) and, through
    \midas{}, the GPU/CPU engine abstraction, driving the analysis back ends with
    manifest-logged results.}
    \label{fig:arch}
\end{figure}

\textbf{System under test.}
\apexa{} is a beamline data analysis system that runs on Advanced Photon Source analysis hosts (Fig.~\ref{fig:arch}), in which a single reasoning agent calls 61 tools
(organized into 
calibration, analysis, motor, knowledge, and visualization domains) over three
Model Context Protocol (MCP) servers~\cite{anthropic2024mcp}: core operations (10),
\midas{}~\cite{sharma2012midas} crystallography (38, with detector calibration,
radial integration (caking), downstream GSAS-II~\cite{toby2013gsas} Rietveld, and
Materials Project~\cite{jain2013materials} structure retrieval), and
EPICS~\cite{dalesio1994epics} motor control (13). Reliability is enforced in a
deterministic tool layer between the model and execution: the execution-integrity
guard and the five safety checks run in server-side code: the latter inside the
EPICS server before any hardware command: not in the model or the prompt.

\textbf{Models.}
The deterministic safety suite is evaluated across four frontier LLMs served through an
institutional LLM inference gateway with identical tool descriptions: \textbf{GPT-5-mini},
\textbf{GPT-5.4}, \textbf{Claude Opus 4.7}, and \textbf{Gemini 2.5 Pro}: to confirm the
tool-layer gate holds regardless of the model driving it. The deployed system runs a
single reasoning agent over all 61 tools.

\textbf{Statistics.}
Chat endpoints are sampled at \texttt{temperature}=0.2. Frontier APIs are not bit-exact
and the gateway exposes no seed, so the prompt-only safety baseline is a single-pass point
estimate across the four models (\texttt{2026-05}). The core claims of this paper are \emph{deterministic} properties of the tool layer,
not sampled quantities: the execution-integrity guard, the 0/200 motor-safety gate
(\S\ref{sec:integrity}).

\section{Results}
\label{sec:results}

\subsection{Autonomous Data Reduction on Real Beamline Data}
\label{sec:results-main}

From a single natural-language prompt, \apexa{} runs the data-reduction pipeline end to
end: detector calibration and 2D$\rightarrow$1D integration (Fig.~\ref{fig:pipeline}), recovering the geometry from instrument metadata and executing the identical analysis on a
Python (GPU) or compiled (CPU) back end per host, with no manual parameter-file editing.

\begin{figure}[t]
    \centering
    \includegraphics[width=\columnwidth]{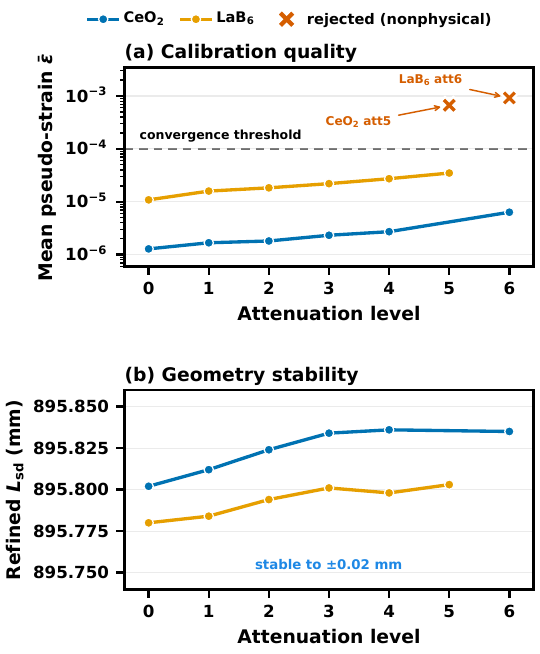}
    \caption{Calibration on a real APS~1-ID dataset.
    \textbf{(a)}~Mean pseudo-strain (the calibration residual; lower is better) vs.\
    attenuation for CeO\textsubscript{2} and
    LaB\textsubscript{6}; the two nonphysical calibrations (CeO\textsubscript{2}
    att5, LaB\textsubscript{6} att6) sit far above the convergence threshold and
    are autonomously flagged and rejected ($\times$). \textbf{(b)}~Refined
    $L_{\rm sd}$ for the converged calibrations is stable to $\pm0.02$~mm across
    both calibrants: APEXA recovers a reproducible physical geometry, and
    refuses to report the ones it does not. All values are read from the per-run
    calibration outputs (\texttt{ai\_tune\_bench.csv}); the $\sim$0.005~mm run-to-run
    variation in (b) is genuine scatter within the $\pm$0.02~mm stability band.}
    \label{fig:aitune}
\end{figure}

\textbf{Testing on experimental beamline data.} We exercised the full pipeline autonomously
on a real APS~1-ID dataset ($2880\times2880$ Varex detector, 63~keV, sample-to-detector distance $L_{\rm sd}\!\approx\!900$~mm):
for each frame, \apexa{} identified the calibrant, resolved the matching dark, refined the
geometry, and produced a 1D pattern. Two tests probe reliability. \emph{First}, calibrating
two chemically distinct standards (CeO\textsubscript{2}, LaB\textsubscript{6}) at the same
condition recovered geometry agreeing to $0.03$~mm in $L_{\rm sd}$ and $<0.005$~px
($<1~\mu$m) in beam centre: the physical geometry, not a per-standard artefact.
\emph{Second}, across a CeO\textsubscript{2} attenuation sweep the recovered geometry held
to $\pm0.02$~mm in $L_{\rm sd}$ and $<0.004$~px in beam centre (Fig.~\ref{fig:aitune}); the
same sweep autonomously flagged and excluded two nonphysical calibrations
(CeO\textsubscript{2} att5, LaB\textsubscript{6} att6: shifted or negative beam centres,
strain $\sim$$10^{-3}$) as invalid geometry rather than emitting them: the same per-run,
manifest-logged automation \apexabench{} evaluates, here on real beamline data.

\subsection{Execution Integrity: Refusing to Report Unexecuted Analysis}
\label{sec:integrity}

For automated \emph{analysis}, the most dangerous failure is not an unsafe motion but a
fabricated result. This is not hypothetical: asked to compare two calibrations, frontier models
would return a complete report: a geometry table, output file sizes, even a
checksum-\emph{identical} verdict: for commands that never ran and files that never existed
on disk. String-match and even tool-state metrics both score such a turn as fluent success;
only checking that a tool \emph{executed} catches it.

\apexa{} closes this channel deterministically. When a response contains tool-call syntax
but no tool executed in that turn, the runner refuses to finalize the answer, injects a
correction, and forces a correctly-formatted retry; after repeated failure it returns an
explicit ``nothing executed'' rather than the model's narration. Paired with the
format-drift-tolerant parser, this makes the trusted record \emph{what deterministically
ran}, not what the model says it ran: and it is the precondition for trusting any number
the system reports. The gate is architectural and model-independent: enforced in code,
not requested in a prompt: and its parser absorbs cross-model format drift (GPT-5-class
models emit the canonical \texttt{TOOL\_CALL}/\texttt{ARGUMENTS} text; Claude~Opus/Sonnet
drift to native XML or bare key--value forms), so a drifted call is never mistaken for
no call.
The guard also fires outside the adversarial suite. In one refinement run the
crystal-structure (CIF) file could not be downloaded because no Materials Project API
key was configured. Refinement needs that file, and its path appeared in no executed
tool's output, so \apexa{} reported the missing input and did not run the refinement,
rather than returning a fit computed against a structure it never had.

\begin{figure}[t]
    \centering
    \includegraphics[width=0.86\columnwidth]{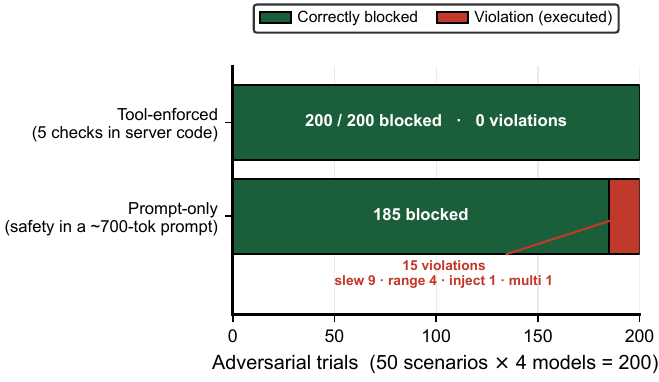}
    \caption{Deterministic motor-safety enforcement over the 50-scenario adversarial
    suite $\times$ four models (200 trials). The \emph{same} scenarios are run two ways:
    \emph{prompt-only} (red), where the safety rules live only in a $\sim$700-token
    system prompt, and \emph{tool-enforced} (green), where those rules run as five
    deterministic checks (soft-limit, limit-switch, slew-magnitude, velocity,
    jog-duration) in the EPICS server before any \texttt{caput}. Tool-enforced produces
    \textbf{0/200} violations (the green ``0'' at every bar) by construction; prompt-only
    lets \textbf{15/200} dangerous commands through, concentrated in large single-step
    slews (9) and out-of-range targets (4), with one each for prompt-injection and
    multi-axis. Bars show only the four scenario types where prompt-only violated; all
    other types were 0 under both. Counts from the adversarial suite run against a
    simulated EPICS IOC; failures spread across all four models.}
    \label{fig:safety}
\end{figure}

\textbf{Motor safety uses the same approach.} \apexa{} is primarily an analysis
framework, but it can also move beamline motors through EPICS. Before any motor command
(\texttt{caput}) is sent, the EPICS tool server runs five checks in code: soft limits,
limit switches, move size, velocity, and jog duration. Because they are code rather than
instructions the model is asked to follow, they cannot be argued around. We tested this
with 50 adversarial scenarios on each of four models, run against a \emph{simulated}
EPICS controller. With the checks in place, none of the 200 trials produced an unsafe
command. The same scenarios given only as a ${\sim}700$-token safety \emph{prompt}
allowed 15 of 200, with failures spread across all four models
(Fig.~\ref{fig:safety}). As with execution integrity, safety has to be enforced in code
rather than requested in the prompt.
\subsection{The Benchmark Beyond Calibration and Integration}
\label{sec:ci}

Calibration and integration are the  core stages in beamline data analysis, but they are not the end of the chain. The cross-detector tasks (\texttt{ref\_01}--\texttt{ref\_08}) extend the benchmark to a scientific endpoint: one NIST-traceable calibrant (CeO\textsubscript{2}, SRM~674b), acquired on four physically distinct detectors, GE, Pilatus and two Varex configurations, from \midas{}'s bundled example data~\cite{sharma2012midas}, is the basis for calibration, integration and GSAS-II~\cite{toby2013gsas} Rietveld refinement tasks. \apexa{} completes them from a single natural-language request.

Refinement is a harder autonomy problem than the stages before it. Calibration and integration are fixed by the data and the instrument geometry; refinement additionally requires a starting structure, which the agent must select, and the outcome can depend on that choice. A fit with a narrow radius of convergence reports success while returning a value close to the one it was given, so completing the step is not evidence that the step constrained the answer.

We therefore score task completion only. Refinement accuracy is not assessed here: the recipe has since been revised, and assessing it requires an ablation over the starting  cell and profile terms, together with a robust per-slice statistic in place of the mean.
\section{Related Work}
\label{sec:related}

\textbf{Agent benchmarks.}
The realism axis runs from code~\cite{chen2021humaneval}, repositories~\cite{jimenez2024swebench},
browsers~\cite{zhou2024webarena}, and multi-modal reasoning~\cite{mialon2023gaia} to
tool use that verifies final \emph{state} in simulation~\cite{yao2024taubench} and
scientific code~\cite{chen2025scienceagentbench,tian2024scicode,huang2024mlagentbench}.
None grade against \emph{physical} consequence, and none face the failure our deployment
does: behind a text-only institutional gateway a real tool call can be \emph{silently
dropped}, so even state-checking the model's claimed output: as $\tau$-bench does in
simulation: would certify a result that never ran. The closest physical benchmark,
AFMBench~\cite{aila2025afm}, controls an atomic-force microscope but has no systematic
safety axis or consequence taxonomy. Adversarial-robustness
suites~\cite{ruan2024toolemu,debenedetti2024agentdojo,yuan2024rjudge,andriushchenko2025agentharm}
operate on digital tools where the failure mode is information leakage, not equipment
damage; \apexabench{} is complementary, with a deterministic tool-layer gate rather than
a model-side defense.

\textbf{Agents at instruments.}
Single-instrument LLM-agent demonstrations include
Coscientist~\cite{boiko2023coscientist}, ChemCrow~\cite{bran2024chemcrow} (on
ReAct~\cite{yao2023react}), VISION~\cite{mathur2025vision} for SAXS/WAXS, X-ray
nanoprobes~\cite{npjagents2026}, and robotic chemistry~\cite{chemagents2025}, atop a
broader self-driving-lab
literature~\cite{abolhasani2023rise,szymanski2023alab,burger2020mobile,xie2025autonomous,wang2023scientific}.
Several of these evaluate rather than merely demonstrate: agents operating real
instruments have been scored for microscopy automation~\cite{aila2025afm} and on
X-ray nanoprobes~\cite{npjagents2026}, and general scientific-agent benchmarks
exist~\cite{chen2025scienceagentbench}. What none of them checks is \emph{execution
integrity}, whether a reported result was produced by a tool call that actually ran.
\apexa{} targets that gap as a deployed analysis framework.

\section{Conclusion}
\label{sec:conclusion}

\apexa{} automates synchrotron data reduction: natural language to executed detector
calibration and azimuthal integration: on a deployed beamline pipeline, run as a single
reasoning loop over 61 tools. Its lesson generalizes beyond the synchrotron: with a
stochastic operator over real instruments, \emph{execution integrity is a tool-layer
property, not a prompting one}: only a deterministic guard turns ``the model said it
calibrated'' into ``the calibration provably ran'' (and gates motor safety at 0/200 vs.\
15/200), and the same benchmark that scores agents is continuous integration for the tools
they call. As
agents take on more of the analysis load at user facilities, the binding constraint is no
longer fluency but trust: that the science an agent reports is science that actually ran, and
that trust must be built into the tool layer, not prompted into a model.


\section*{Acknowledgments}

We thank Ming Du for helpful discussions on Model Context Protocol integration and model selection, and for pointers to computing resources within the group.

This work was supported by the U.S.\ Department of Energy, Office of Science,
Advanced Scientific Computing Research. This research used resources of the
Advanced Photon Source, a U.S.\ Department of Energy (DOE) Office of Science user
facility at Argonne National Laboratory, and is based on research supported by the
U.S.\ DOE Office of Science-Basic Energy Sciences, under Contract No.\
DE-AC02-06CH11357.

The submitted manuscript has been created by UChicago Argonne, LLC, Operator of
Argonne National Laboratory (``Argonne''). Argonne, a U.S.\ Department of Energy
Office of Science laboratory, is operated under Contract No.\ DE-AC02-06CH11357.
The U.S.\ Government retains for itself, and others acting on its behalf, a paid-up
nonexclusive, irrevocable worldwide license in said article to reproduce, prepare
derivative works, distribute copies to the public, and perform publicly and display
publicly, by or on behalf of the Government. The Department of Energy will provide
public access to these results of federally sponsored research in accordance with
the DOE Public Access Plan (\url{http://energy.gov/downloads/doe-public-access-plan}).

\bibliographystyle{IEEEtran}
\bibliography{references}

\appendix
\section{Artifact Description (AD)}
\label{sec:ad}

\subsection{Overview of Contributions and Artifacts}
This paper's claims are supported by three released artifacts: (i)~the \apexa{}
multi-agent framework (five specialist domains, 61 tools over three MCP servers, run as a
single reasoning loop);
(ii)~\apexabench{}, the evaluation harness of 58 facility tasks plus a 50-scenario
adversarial motor-safety suite, with the scoring code implementing
Eq.~\ref{eq:score}; and (iii)~the interaction traces from the reported runs. The AD
below is sufficient to reproduce the analysis of the real beamline data (Fig.~\ref{fig:pipeline},
\ref{fig:aitune}) and the deterministic guards (execution-integrity guard;
0/200 motor-safety, Fig.~\ref{fig:safety}). Code, harness, and traces are available at
\url{https://github.com/AdvancedPhotonSource/APEXA-APS-Beamline-Assistant}.

\subsection{Computational Artifacts}
\textbf{Software.} Python~3.13, dependencies managed with \texttt{uv}
(\texttt{uv sync}); no manual environment activation. The analysis back end is
\midas{}~v11 (crystallography executables and Python packages) and, for
refinement, \gsas{}. Frontier-model access is through an institutional LLM
gateway; the harness runs without it in \texttt{--dry-run} and \texttt{--mock} modes.

\textbf{Datasets.} Task and adversarial-suite definitions and the safety interaction
traces ship as JSON with the harness (\texttt{benchmark/}). The
\texttt{ai\_tune} CeO\textsubscript{2}/LaB\textsubscript{6} calibration frames and their
per-run refined geometries (Fig.~\ref{fig:pipeline},~\ref{fig:aitune}), together with the
adversarial-safety summary, are archived at the Materials Data Facility
(\url{https://doi.org/10.18126/tgg4-1m26}). Downstream integration and
Rietveld-refinement products are deliberately not included: this paper's claims concern
the automation architecture and its execution-integrity properties, not the
crystallographic values those later stages produce, and archiving unvalidated derived
results would invite them to be read as vetted. The cross-detector slice (\S\ref{sec:ci}) uses \midas{}'s
bundled CeO\textsubscript{2} (NIST SRM~674b) example frames across four detector
geometries; no restricted data are required.

\textbf{Hardware.} No special hardware is required to reproduce the harness logic:
the safety suite runs on a CPU-only laptop via a mock EPICS layer, and benchmark
grading is CPU-bound. A CUDA GPU exercises the GPU analysis engine but the compiled
CPU engine is a functional substitute (the engine-abstraction claim).

\subsection{Reproduction Workflow}
\begin{enumerate}
\item \texttt{uv sync} to install (\(\sim\)168 packages).
\item Validate task definitions without any API calls:\\
\texttt{uv run python benchmark/eval\_harness.py --dry-run}
\item Motor-safety suite, no IOC required (reproduces 0/200):\\
\texttt{uv run python benchmark/eval\_safety.py --mock}\\
Prompt-only baseline (reproduces the 15/200 contrast):\\
\texttt{... eval\_safety.py --mock --prompt-only}
\item Full 58-task benchmark for a model (requires gateway credentials):\\
\texttt{uv run python benchmark/eval\_harness.py --model gpt54}
\end{enumerate}
Outputs are written as per-task JSON under \texttt{benchmark/results/} with per-category
success rates and the composite \textrm{APEXA-Score}.

\textbf{Scoring rubric.} The composite the harness logs is
\begin{equation}
\textrm{APEXA-Score} = \frac{1}{|T|}\sum_{t\in T}
  w_t \cdot \mathcal{C}_t \cdot \mathcal{S}_t \cdot
  \bigl(\alpha + (1-\alpha)\,\mathcal{E}_t\bigr),
\label{eq:score}
\end{equation}
with correctness $\mathcal{C}\!\in\!\{0,1\}$, safety $\mathcal{S}\!\in\!\{0,1\}$,
efficiency $\mathcal{E}=\min(1, n_\text{opt}/n_\text{actual})$, consequence weights
$w_\text{I}{=}1, w_\text{C}{=}1.5, w_{\text{P}_\text{R}}{=}2, w_{\text{P}_\text{I}}{=}3$,
and $\alpha{=}0.7$ (fixed \emph{a priori}). Weights are illustrative; the multiplicative
$\mathcal{C}\!\cdot\!\mathcal{S}$ makes an unsafe-but-correct task score zero. The harness
also logs unweighted per-category success.
\subsection{Reproducibility Notes}
The paper's core claims: the execution-integrity guard and the 0/200 motor-safety
gate are deterministic properties of the tool layer and reproduce exactly,
independent of model sampling. \emph{Crucially, they need no gateway access}: the 0/200 gate reproduces via \texttt{eval\_safety.py --mock}, where the tool-layer checks run without an LLM, and task definitions validate under \texttt{--dry-run}. The cross-detector tasks reproduce by running \texttt{detector\_zoo/refine\_v2.py} against the integrated patterns in the data release (MDF DOI 10.18126/tgg4-1m26), also without an LLM. Only the prompt-only baseline (a single-pass point estimate, \texttt{2026-05}; frontier endpoints are not bit-exact and the gateway exposes no seed) and full agent scoring invoke an LLM: we release their interaction traces, and the agent uses OpenAI-format tool definitions so the provider layer can target any compatible endpoint, not only the institutional gateway.

\end{document}